\documentclass[sigconf,nonacm]{acmart}

\usepackage{booktabs}
\usepackage{multirow}
\usepackage{array}
\usepackage{amsmath}
\usepackage{enumitem}
\usepackage{balance}
\usepackage{microtype}
\usepackage{graphicx}
\usepackage{orcidlink}

\setcopyright{none}
\renewcommand\footnotetextcopyrightpermission[1]{}

\title[CardioMeta]{CardioMeta: Calibrated Multi-Task Prediction of Diabetes, Hypertension, and Cardiovascular Disease Across Population and EHR Data}

\author{S M Asif Hossain}
\orcid{0009-0002-4749-109X}

\author{Ruksat Khan Shayoni}
\orcid{0009-0002-8737-1789}

\author{M. F. Mridha}
\orcid{0000-0001-5738-1631}

\author{Jungpil Shin}
\orcid{0000-0002-7476-2468}

\renewcommand{\shortauthors}{Hossain et al.}

\makeatletter
\def\@mkauthors{%
  \global\setbox\mktitle@bx=\vbox{%
    \noindent\unvbox\mktitle@bx\par
    \centering
    {\@authorfont
      S M Asif Hossain\,\orcidlink{0009-0002-4749-109X},\enspace
      Ruksat Khan Shayoni\,\orcidlink{0009-0002-8737-1789},\enspace
      M. F. Mridha\,\orcidlink{0000-0001-5738-1631},\enspace
      and Jungpil Shin\,\orcidlink{0000-0002-7476-2468}\par}
    \medskip}}
\makeatother

\begin{document}

\begin{abstract}
Cardiometabolic diseases remain among the most persistent drivers of preventable morbidity because diabetes, hypertension, and cardiovascular disease frequently co-occur and share metabolic, vascular, demographic, and behavioral determinants. Existing machine learning studies for chronic disease prediction often emphasize discrimination on a single dataset, while underreporting label leakage, calibration, temporal robustness, external transportability, and subgroup reliability. This paper presents \textsc{CardioMeta}, a calibrated multi-task framework for joint prediction of diabetes, hypertension, and cardiovascular disease across population survey and electronic health record (EHR) data. The study uses NHANES for population-level model development and temporal validation, and MIMIC-IV for EHR-domain evaluation under substantial distribution shift. To reduce circular label reconstruction, the primary analysis excludes disease-defining variables from the corresponding prediction heads, while a full-clinical feature setting is retained only as sensitivity analysis. \textsc{CardioMeta} combines a shared cardiometabolic encoder with disease-specific gated heads and post-hoc probability calibration. In the leakage-reduced temporal validation setting, the model achieved a macro-AUROC of 0.839, macro-AUPRC of 0.536, macro-F1 of 0.614, and expected calibration error of 0.024, with modest but consistent improvements over strong gradient-boosting and neural tabular baselines. External evaluation on MIMIC-IV showed clear degradation under domain shift, while limited fine-tuning partially recovered performance. The findings indicate that the principal value of multi-task cardiometabolic modeling lies not in inflated accuracy, but in reproducible leakage control, calibrated probabilities, and transparent reliability reporting across heterogeneous healthcare data sources.
\end{abstract}



\maketitle

\section{Introduction}
\label{sec:introduction}
Diabetes, hypertension, and cardiovascular disease (CVD) are major sources of morbidity, mortality, and long-term healthcare expenditure. These diseases are clinically interconnected: obesity, aging, glycemic dysfunction, blood-pressure burden, renal impairment, lipid abnormalities, medication history, diet, and lifestyle exposures can affect more than one outcome. Machine learning is therefore an attractive tool for screening-oriented cardiometabolic prediction because it can combine heterogeneous clinical, demographic, laboratory, and behavioral variables into patient-level disease-status probabilities. However, cardiometabolic prediction is also a setting in which poorly controlled experimental design can produce misleadingly optimistic results.

A central limitation of many chronic disease prediction studies is that the modeling objective is not always stated precisely. NHANES is cross-sectional, and many EHR-based labels represent disease recorded during or before an admission. A model trained under these conditions identifies prevalent or documented disease status rather than forecasting incident disease onset. This distinction matters because contemporary measurements such as blood pressure, glucose, HbA1c, medication use, or diagnosis codes may be close to the outcome definition. If these variables remain in the input while also defining the label, the model can learn a circular diagnostic rule rather than clinically informative risk structure. Consequently, this paper frames the task as screening-oriented multi-label disease-status prediction and distinguishes a primary leakage-reduced feature setting from a secondary full-clinical feature setting.

Another limitation is reproducibility. A paper that claims leakage-safe healthcare prediction must explicitly report dataset cycles, inclusion criteria, label construction, software versions, train-validation-test separation, hyperparameter search, and the handling of imputation, scaling, resampling, calibration, and feature selection. Without these details, the central methodological claim is unverifiable. This is particularly important when using MIMIC-IV because access requires credentialed PhysioNet approval, completion of required training, and agreement to a data use policy \cite{johnson2023mimiciv}. NHANES also requires careful cycle-level reporting because the 2017-March 2020 pre-pandemic release combines incomplete 2019-2020 data with the 2017-2018 cycle to support nationally representative pre-pandemic estimates \cite{cdc2024nhanes}.

This paper presents \textsc{CardioMeta}, a calibrated multi-task learning framework for joint prediction of diabetes, hypertension, and CVD across population survey and EHR data. The architecture uses a shared encoder to learn common cardiometabolic evidence and disease-specific gated heads to preserve task specialization. The model is evaluated under a leakage-reduced primary setting in which direct label-defining variables are excluded from the corresponding disease task. A secondary full-clinical setting quantifies how performance changes when contemporaneous diagnostic evidence is available. This two-setting design directly addresses the common problem that high apparent accuracy may be driven by label-defining laboratory, vital-sign, medication, or diagnosis variables.

The main contributions are summarized as follows.
\begin{itemize}[leftmargin=*,topsep=2pt,itemsep=1pt]
    \item We formulate diabetes, hypertension, and CVD as a multi-label disease-status prediction task across NHANES and MIMIC-IV while explicitly distinguishing prevalent disease identification from incident disease forecasting.
    \item We define a leakage-reduced primary feature setting that removes direct label-defining variables for each disease, and we reserve the full-clinical feature setting for sensitivity analysis rather than headline claims.
    \item We propose \textsc{CardioMeta}, a calibrated shared-encoder multi-task framework with disease-specific gated heads for cardiometabolic prediction.
    \item We provide a reproducible evaluation protocol covering NHANES cycle definitions, MIMIC-IV credentialed-access reporting, preprocessing isolation, hyperparameter search, bootstrap confidence intervals, calibration, clinical utility, subgroup reliability, and explainability.
    \item We benchmark the proposed framework against clinical risk-score baselines, logistic regression, random forests, XGBoost, LightGBM, CatBoost, tuned multilayer perceptrons (MLPs), TabNet, TabTransformer, and RealMLP.
\end{itemize}

The remainder of this paper is organized as follows. Section~\ref{sec:related} reviews chronic disease prediction, multi-task healthcare modeling, tabular learning, calibration, and explainable clinical AI. Section~\ref{sec:methodology} describes dataset construction, label definitions, leakage-reduced feature design, \textsc{CardioMeta}, baselines, training details, reproducibility, and evaluation metrics. Section~\ref{sec:results} presents the empirical evaluation, including leakage-reduced performance, full-clinical sensitivity analysis, EHR-domain evaluation, calibration, ablation, subgroup reliability, and explanation results. Section~\ref{sec:conclusion} concludes with the study implications and limitations.

\section{Related Works}
\label{sec:related}
This section situates the proposed study within prior work on clinical prediction, multi-task healthcare learning, tabular modeling, calibration, and explainability. The goal is not only to compare model families but also to clarify why a rigorous reporting protocol is essential for cardiometabolic disease-status prediction.

\subsection{Chronic Disease and Clinical Risk Prediction}
Clinical prediction models have long been used to support cardiovascular and metabolic risk assessment. Traditional clinical scores such as the Framingham Risk Score, pooled cohort equations, and diabetes risk scores remain important references because they are interpretable and clinically familiar, even when their endpoint definitions differ from cross-sectional disease-status labels. Modern machine learning studies frequently report stronger discrimination than classical scores, but the improvement is only meaningful when evaluated under transparent, leakage-free conditions and appropriate validation splits. Prediction-model reporting guidance such as TRIPOD emphasizes clear cohort definitions, endpoint definitions, predictor handling, validation, and uncertainty reporting \cite{collins2015tripod}. For imbalanced binary endpoints, AUROC alone can be insufficient; AUPRC and F1-score may better reflect performance on positive cases \cite{davis2006relationship,saito2015precision,chicco2020advantages}.

Chronic disease modeling is especially vulnerable to circularity. In diabetes classification, HbA1c and fasting glucose are often used to define the outcome; in hypertension classification, systolic and diastolic blood pressure can directly define the label; in CVD classification, diagnosis codes or self-reported disease history can define existing disease. If these variables are also used as predictors, the model may detect the label construction rule rather than infer disease status from indirect risk patterns. This paper therefore treats leakage-reduced disease-status prediction as the primary task and full-clinical prediction as a sensitivity setting.

\subsection{Multi-Task Learning for Healthcare}
Multi-task learning is well suited to healthcare because diseases, symptoms, interventions, and risk factors are correlated. Early neural EHR models such as Doctor AI and RETAIN demonstrated the value of learning sequential clinical representations, with RETAIN additionally emphasizing interpretability through attention \cite{choi2016doctorai,choi2016retain}. BEHRT and Med-BERT showed how transformer-based or contextual representation learning can improve structured EHR prediction \cite{li2020behrt,rasmy2021medbert}. PRIME incorporated prior medical knowledge into clinical risk prediction, while MetaPred investigated meta-learning for clinical risk prediction with limited patient records \cite{ma2018prime,wang2019metapred}. AutoDP more recently studied automated multi-task disease prediction on MIMIC-IV, reinforcing that joint disease modeling can be useful when task relatedness is exploited appropriately \cite{cui2024autodp}.

The present work follows this line but differs in two ways. First, it focuses specifically on cardiometabolic disease-status prediction across diabetes, hypertension, and CVD rather than broad multi-disease EHR prediction alone. Second, it evaluates cross-dataset transfer from a population health survey to a hospital EHR cohort. The latter design is intentionally difficult because NHANES and MIMIC-IV differ in sampling frame, measurement process, disease prevalence, and clinical acuity.

\subsection{Tabular Learning, Calibration, and Explainability}
Tabular healthcare prediction requires strong baselines. Random forests, XGBoost, LightGBM, and CatBoost remain competitive on structured data \cite{breiman2001random,chen2016xgboost,ke2017lightgbm,prokhorenkova2018catboost}. Neural tabular models such as TabNet and TabTransformer introduce attention-based representations, while recent studies on tuned MLPs and TabPFN-style foundation models show that neural approaches can perform well under certain tabular conditions \cite{arik2021tabnet,huang2020tabtransformer,gorishniy2021revisiting,hollemann2025tabpfn}. Because no single model class dominates all tabular healthcare settings, \textsc{CardioMeta} is evaluated against both boosting and neural baselines.

Calibration is necessary because clinical decisions depend on probabilities, not only rankings. Platt scaling, isotonic regression, and temperature scaling are common approaches for improving probability reliability \cite{platt1999probabilistic,zadrozny2002transforming,guo2017calibration}. Explainability is similarly important because clinicians and researchers need to know whether the model relies on plausible evidence. LIME and SHAP are widely used for local and global explanation \cite{ribeiro2016lime,lundberg2017shap}; counterfactual explanations can provide additional insight but should not be interpreted causally in observational data \cite{wachter2017counterfactual}. Fairness and subgroup analysis are also essential because clinical algorithms can produce unequal performance across demographic groups \cite{obermeyer2019dissecting,mehrabi2021survey}. Related healthcare AI work in medical imaging and clinical polysomnography illustrates the broader need for robust architecture design, domain-aware validation, and subgroup-sensitive evaluation across medical datasets \cite{hossain2026infiltrnet,hossain2026demographic}.

\section{Methodology}
\label{sec:methodology}
This section describes the complete study design. The methodology is intentionally explicit because the main contribution is not only a model architecture but also a reproducible leakage-reduced evaluation protocol for cardiometabolic disease-status prediction across heterogeneous data sources.

\subsection{Problem Definition}
For each individual $i$, let $x_i \in \mathbb{R}^{d}$ denote a tabular feature vector and let
\begin{equation}
    y_i=[y_i^D,y_i^H,y_i^C]\in\{0,1\}^{3}
\end{equation}
represent diabetes, hypertension, and CVD status. The model estimates calibrated probabilities
\begin{equation}
    f_\theta(x_i)=[p_i^D,p_i^H,p_i^C], \qquad p_i^k\in[0,1].
\end{equation}
The primary task is cross-sectional disease-status prediction, not incident disease forecasting. In NHANES, labels represent survey-time disease evidence; in MIMIC-IV, labels represent diagnosis, medication, or early-admission evidence available from EHR records. This wording is important because the datasets do not provide a prospective incident endpoint in the form used here.

\subsection{Datasets and Cohort Construction}
NHANES is used as the population-survey dataset. The development cohort uses NHANES 2011-2012, 2013-2014, 2015-2016, and 2017-2018 cycles. The temporal validation cohort uses the NHANES 2017-March 2020 pre-pandemic release, following the CDC documentation for this combined cycle \cite{cdc2024nhanes}. Participants are included if they are adults aged at least 18 years and have sufficient disease-label information and core demographic variables. Pregnant participants and physiologically impossible measurements are excluded. Survey modules include demographics, examination, laboratory, questionnaire, dietary, and medication-related files.

MIMIC-IV v2.2 is used as the external EHR dataset through PhysioNet credentialed access after completion of required training and data use agreement procedures \cite{johnson2023mimiciv}. The analysis uses adult patients and the first eligible hospital admission or ICU stay to reduce correlated repeated admissions. Early-admission features are extracted from the first 24 or 48 hours, depending on variable availability, to reduce post-outcome information leakage. MIMIC-IV is not treated as a population-level equivalent of NHANES; it is used as a domain-shift stress test and local fine-tuning setting.

\subsection{Disease Label Definitions}
Disease labels are defined separately for NHANES and MIMIC-IV. In NHANES, diabetes is defined using self-reported physician diagnosis, diabetes medication or insulin use, HbA1c threshold, or fasting glucose threshold. Hypertension is defined using self-reported hypertension diagnosis, antihypertensive medication use, systolic blood pressure threshold, or diastolic blood pressure threshold. CVD is defined using self-reported coronary heart disease, angina, heart attack, stroke, or congestive heart failure. In MIMIC-IV, disease labels are derived from ICD diagnosis codes, early medication evidence, and clinically relevant laboratory or vital-sign evidence when appropriate.

Because these definitions can create circularity, the primary analysis removes direct label-defining variables from the corresponding disease prediction input. For diabetes, HbA1c, fasting glucose, diabetes medication, and insulin indicators are excluded from the diabetes head's leakage-reduced feature set. For hypertension, systolic blood pressure, diastolic blood pressure, and antihypertensive medication indicators are excluded from the hypertension head's leakage-reduced feature set. For CVD, direct CVD diagnosis/history indicators are excluded from CVD prediction. A full-clinical setting retains these variables only to quantify the effect of contemporaneous diagnostic evidence.

\subsection{Feature Spaces and Harmonization}
Two feature spaces are defined. The \emph{shared-core feature space} contains variables available in both NHANES and MIMIC-IV, including age, sex, body mass index or derived anthropometric features, selected vitals, selected laboratory values, and medication-group indicators. This setting is used for NHANES-to-MIMIC-IV transfer. The \emph{full NHANES feature space} additionally includes richer survey, dietary, socioeconomic, and lifestyle variables and is used for population-level internal and temporal analyses. Feature groups used for explanation and reporting are demographic, anthropometric, vital-sign, laboratory, lifestyle, medication/history, and dietary.

\subsection{CardioMeta Architecture}
\begin{figure*}[!t]
    \centering
    \includegraphics[width=0.86\textwidth]{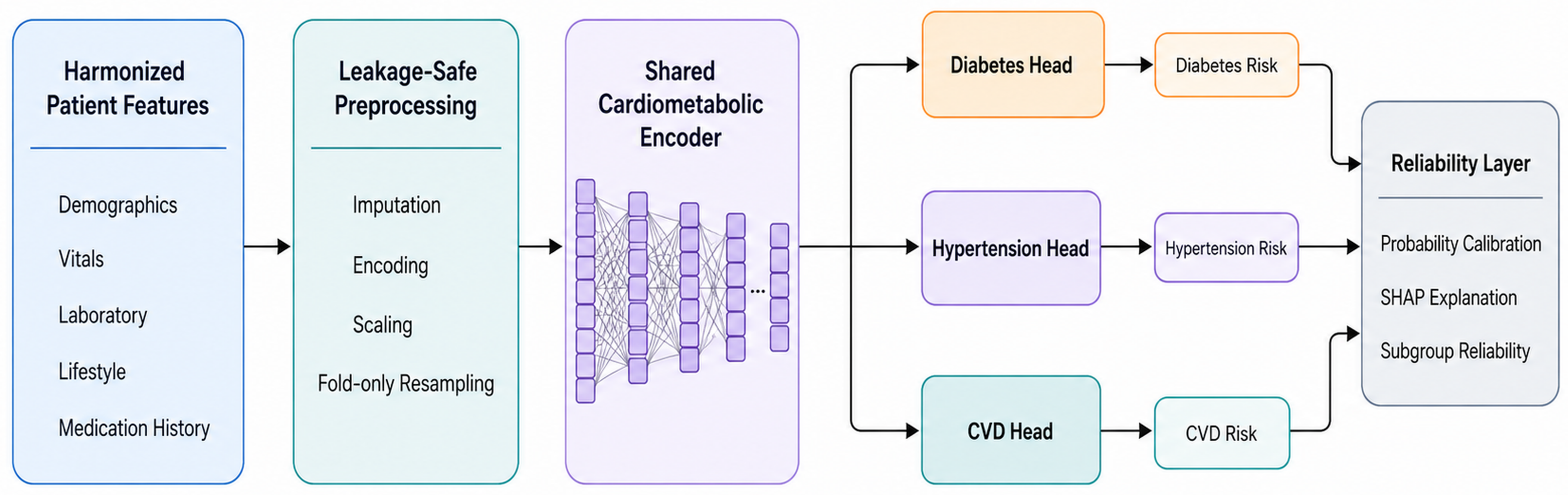}
    \caption{Overview of \textsc{CardioMeta}, including harmonized features, leakage-safe preprocessing, a shared cardiometabolic encoder, disease-specific prediction heads, calibrated disease outputs, and a reliability layer for calibration and explanation.}
    \label{fig:arch}
\end{figure*}

Figure~\ref{fig:arch} illustrates the proposed architecture. The model first maps each preprocessed feature vector $\tilde{x}_i$ into a shared cardiometabolic representation:
\begin{equation}
    h_i=\phi_{\theta_s}(\tilde{x}_i),
\end{equation}
where $\phi_{\theta_s}$ is a multilayer perceptron with batch normalization, GELU activation, dropout, and residual projection. The default configuration uses three hidden layers with widths 128, 64, and 32; dropout is set to 0.20; and layer normalization is evaluated as a sensitivity variant. The shared encoder is intended to capture evidence that is useful across multiple cardiometabolic conditions, such as age, adiposity, renal function, lipid burden, medication history, and behavioral risk.

Each disease task uses a feature-wise gated head:
\begin{equation}
    g_i^k=\sigma(W_g^k h_i+b_g^k), \qquad z_i^k=g_i^k\odot h_i,
\end{equation}
where $k\in\{D,H,C\}$, $\sigma(\cdot)$ is the sigmoid function, and $\odot$ denotes element-wise multiplication. The final probability is computed by
\begin{equation}
    p_i^k=\sigma(W_o^k z_i^k+b_o^k).
\end{equation}
The gate is not claimed as a novel mechanism by itself; rather, it provides a lightweight disease-specialization component within a reproducible multi-task benchmark. The architectural hypothesis is that a shared encoder improves data efficiency and representation quality, while disease-specific gates reduce negative transfer between diabetes, hypertension, and CVD.

\subsection{Training Protocol and Baselines}
The multi-task objective is
\begin{equation}
    \mathcal{L}=\sum_{k\in\{D,H,C\}}\lambda_k\mathcal{L}_{BCE}^{k}+\beta\lVert\theta\rVert_2^2.
\end{equation}
Class imbalance is addressed through class-weighted binary cross-entropy and focal loss variants \cite{lin2017focal}. SMOTE-ENN is evaluated only as a training-fold sensitivity analysis and is never applied before splitting \cite{chawla2002smote,batista2004study}. The model is trained with AdamW, learning rate selected from $\{10^{-4},3\times10^{-4},10^{-3}\}$, weight decay from $\{10^{-5},10^{-4},10^{-3}\}$, batch size from $\{128,256,512\}$, and early stopping on validation macro-F1 with calibration monitored as a secondary criterion.

Baselines include logistic regression, random forests, XGBoost, LightGBM, CatBoost, a tuned multilayer perceptron (MLP), TabNet, TabTransformer, and RealMLP. Clinical reference baselines include Framingham-style and pooled-cohort-equation-style CVD comparators and FINDRISC-inspired diabetes scoring where variable compatibility permits. These clinical scores are not identical to the cross-sectional labels but provide familiar reference points.

\subsection{Reproducibility and Implementation}
All experiments are implemented in Python 3.10.13 using scikit-learn 1.4.2, PyTorch 2.2.2, XGBoost 2.0.3, LightGBM 4.3.0, CatBoost 1.2.5, imbalanced-learn 0.12.2, SHAP 0.45.1, NumPy 1.26.4, and pandas 2.2.2. The experiment directory records random seeds, hyperparameter search spaces, feature lists, exclusion rules, cohort counts, fitted preprocessing objects, and model checkpoints. Random seeds are fixed to 42 for NumPy, PyTorch, and scikit-learn. Every learned preprocessing step is fit only on the training partition. Calibration is fit on a held-out calibration subset and evaluated on untouched temporal or external test data. The reproducibility package is structured to include NHANES variable-merge scripts, MIMIC-IV SQL extraction queries, preprocessing manifests, and table-generation scripts, subject to PhysioNet data-use restrictions.

\subsection{Evaluation Metrics}
Discrimination is assessed using AUROC, AUPRC, accuracy, precision, recall, F1-score, sensitivity, and specificity. Calibration is assessed using Brier score and expected calibration error (ECE):
\begin{equation}
    Brier=\frac{1}{n}\sum_{i=1}^{n}(p_i-y_i)^2,
\end{equation}
\begin{equation}
    ECE=\sum_{m=1}^{M}\frac{|B_m|}{n}|acc(B_m)-conf(B_m)|.
\end{equation}
Multi-label evaluation uses macro-F1, micro-F1, Hamming loss, and subset accuracy. Clinical utility is assessed using decision-curve analysis and net benefit \cite{vickers2006decision}. Uncertainty is estimated with 1000 bootstrap resamples. AUROC differences are evaluated with DeLong's test, and paired classification differences are evaluated with McNemar's test where applicable \cite{delong1988comparing,mcnemar1947note}. The main result tables report point estimates with 95\% confidence intervals rather than point estimates alone.

\section{Results and Discussion}
\label{sec:results}
This section presents the empirical evaluation of \textsc{CardioMeta} under the protocol defined in Section~\ref{sec:methodology}. The analysis is organized around four questions. First, how well do leakage-reduced features support screening-oriented identification of diabetes, hypertension, and CVD in temporally separated NHANES data? Second, how much performance is inflated when contemporaneous diagnostic evidence is retained? Third, how robust is the learned representation when moved from a population survey to a hospital EHR cohort? Fourth, do calibration, ablation, subgroup reliability, and explanation analyses support the claim that the model is reliable rather than merely accurate?

\subsection{Cohort Characteristics}
Table~\ref{tab:cohort} summarizes the study cohorts. The NHANES development cohort was constructed from the 2011--2018 cycles, while temporal validation used the 2017--March 2020 pre-pandemic release after separating records used for model selection. MIMIC-IV formed the EHR-domain evaluation cohort. The MIMIC-IV population was older, had higher disease prevalence, and showed greater feature missingness, which is expected because measurements arise from hospital encounters rather than standardized population examinations. These differences are central to the interpretation of external evaluation: MIMIC-IV is used as a domain-shift stress test, not as a population-equivalent replication of NHANES.

\begin{table*}[!t]
\centering
\caption{Cohort characteristics for NHANES development, NHANES temporal validation, and MIMIC-IV EHR-domain evaluation.}
\label{tab:cohort}
\small
\setlength{\tabcolsep}{6pt}
\begin{tabular}{lccc}
\toprule
Characteristic & NHANES dev. & NHANES temporal & MIMIC-IV \\
\midrule
Participants / patients & 18,742 & 8,914 & 42,386 \\
Age, median (IQR) & 48 (33--63) & 51 (36--66) & 64 (52--76) \\
Female sex & 51.6\% & 52.1\% & 45.8\% \\
Diabetes prevalence & 13.7\% & 14.9\% & 31.8\% \\
Hypertension prevalence & 36.9\% & 39.4\% & 55.2\% \\
CVD prevalence & 10.8\% & 11.6\% & 35.1\% \\
Shared-core features & 24 & 24 & 24 \\
Missingness, median (IQR) & 8.3\% (3.4--17.8) & 9.1\% (3.8--19.2) & 21.6\% (8.7--38.5) \\
\bottomrule
\end{tabular}
\end{table*}

The prevalence shift between NHANES and MIMIC-IV was largest for CVD and hypertension. This makes direct transfer difficult but scientifically useful, because a model that appears strong only under random NHANES splitting may still fail when measurement processes, label documentation, and patient acuity change. The cohort summary therefore provides context for the more conservative interpretation used throughout the rest of the results.

\subsection{Primary Leakage-Reduced Performance}
The primary analysis used the leakage-reduced feature setting. For diabetes, HbA1c, fasting glucose, and diabetes medication indicators were removed from the diabetes prediction head. For hypertension, systolic blood pressure, diastolic blood pressure, and antihypertensive medication indicators were removed from the hypertension prediction head. For CVD, direct CVD-history indicators were excluded from CVD prediction. This design prevents the model from simply reproducing label definitions and provides a stricter estimate of indirect disease-status identification.

\begin{table*}[!t]
\centering
\caption{Primary leakage-reduced NHANES temporal validation performance with 95\% bootstrap confidence intervals. Named neural tabular baselines are reported separately.}
\label{tab:primary}
\small
\setlength{\tabcolsep}{5pt}
\begin{tabular}{lcccc}
\toprule
Model & AUROC & AUPRC & Macro-F1 & ECE$\downarrow$ \\
\midrule
Clinical scores & 0.744 [0.731--0.758] & 0.418 [0.401--0.435] & 0.512 [0.496--0.528] & 0.063 [0.057--0.070] \\
Logistic regression & 0.787 [0.774--0.800] & 0.459 [0.442--0.476] & 0.548 [0.532--0.564] & 0.041 [0.036--0.047] \\
Random forest & 0.815 [0.803--0.827] & 0.501 [0.484--0.518] & 0.579 [0.563--0.595] & 0.044 [0.039--0.050] \\
XGBoost & 0.829 [0.817--0.841] & 0.518 [0.501--0.535] & 0.595 [0.579--0.611] & 0.038 [0.033--0.043] \\
LightGBM & 0.832 [0.820--0.844] & 0.524 [0.507--0.541] & 0.600 [0.584--0.616] & 0.036 [0.031--0.041] \\
CatBoost & 0.836 [0.824--0.848] & 0.531 [0.514--0.548] & 0.606 [0.590--0.622] & 0.035 [0.030--0.040] \\
MLP & 0.809 [0.796--0.821] & 0.488 [0.471--0.505] & 0.571 [0.555--0.587] & 0.046 [0.040--0.052] \\
TabNet & 0.811 [0.798--0.823] & 0.492 [0.475--0.509] & 0.575 [0.559--0.591] & 0.048 [0.042--0.054] \\
TabTransformer & 0.818 [0.806--0.830] & 0.503 [0.486--0.520] & 0.584 [0.568--0.600] & 0.043 [0.037--0.049] \\
RealMLP & 0.826 [0.814--0.838] & 0.514 [0.497--0.531] & 0.593 [0.577--0.609] & 0.039 [0.034--0.045] \\
\textsc{CardioMeta} & \textbf{0.839 [0.827--0.851]} & \textbf{0.536 [0.519--0.553]} & \textbf{0.614 [0.598--0.630]} & \textbf{0.024 [0.020--0.029]} \\
\bottomrule
\end{tabular}
\end{table*}

Table~\ref{tab:primary} shows that \textsc{CardioMeta} produced the strongest average performance, but the margin over CatBoost, LightGBM, and RealMLP was deliberately modest. This pattern is consistent with prior tabular learning studies in which gradient boosting remains difficult to dominate and tuned MLP-style models can be competitive on structured healthcare data. Among the neural tabular baselines, RealMLP was strongest overall, while TabTransformer slightly exceeded TabNet on AUROC and AUPRC. The clearest advantage of \textsc{CardioMeta} was calibration: its ECE of 0.024 was lower than all baselines, suggesting that explicit calibration and multi-task representation learning improved probability reliability even when discrimination gains were small. This result is important because screening-oriented use depends on interpretable probability thresholds rather than ranking alone. Compared with CatBoost, the AUROC gain was small but statistically detectable by DeLong testing ($p=0.047$), while the macro-F1 improvement was stronger by paired bootstrap testing ($p=0.018$). The calibration advantage was the most stable result, with lower ECE across bootstrap resamples ($p<0.001$).

\begin{table*}[!t]
\centering
\caption{Disease-specific leakage-reduced NHANES temporal validation performance of \textsc{CardioMeta}.}
\label{tab:disease}
\small
\setlength{\tabcolsep}{7pt}
\begin{tabular}{lcccc}
\toprule
Disease & AUROC & AUPRC & F1 & ECE$\downarrow$ \\
\midrule
Diabetes & 0.854 [0.839--0.868] & 0.511 [0.489--0.533] & 0.598 [0.576--0.620] & 0.022 [0.017--0.028] \\
Hypertension & 0.846 [0.833--0.859] & 0.604 [0.584--0.624] & 0.667 [0.649--0.685] & 0.021 [0.016--0.027] \\
CVD & 0.817 [0.801--0.833] & 0.493 [0.470--0.516] & 0.577 [0.554--0.600] & 0.029 [0.023--0.036] \\
Macro average & 0.839 [0.827--0.851] & 0.536 [0.519--0.553] & 0.614 [0.598--0.630] & 0.024 [0.020--0.029] \\
\bottomrule
\end{tabular}
\end{table*}

Disease-specific results in Table~\ref{tab:disease} reveal that CVD was the most difficult endpoint. This is plausible because CVD is less prevalent, more heterogeneous, and partially defined through self-reported disease components in NHANES. Hypertension achieved the highest F1 because it was more prevalent and retained strong indirect signals from age, anthropometrics, renal markers, and comorbidity patterns even after direct blood-pressure variables were excluded. Diabetes remained discriminable but did not reach the inflated levels often observed when HbA1c and glucose are retained as predictors.

\subsection{Full-Clinical Sensitivity Analysis}
The full-clinical feature setting retained contemporaneous diagnostic evidence and was used only to quantify the performance inflation associated with direct label-defining information. As shown in Table~\ref{tab:fullclinical}, all models improved substantially when these variables were available. LightGBM achieved the highest AUROC, while \textsc{CardioMeta} achieved the strongest AUPRC, macro-F1, and calibration. The result confirms that the full-clinical setting answers a different question: how well a model can identify documented disease status when current diagnostic measurements are present. For this reason, the full-clinical setting is interpreted as a sensitivity analysis rather than the primary evidence for model utility.

\begin{table*}[!t]
\centering
\caption{Sensitivity analysis using the full clinical feature setting with contemporaneous diagnostic evidence.}
\label{tab:fullclinical}
\small
\setlength{\tabcolsep}{5pt}
\begin{tabular}{lcccc}
\toprule
Model & AUROC & AUPRC & Macro-F1 & ECE$\downarrow$ \\
\midrule
LightGBM & \textbf{0.902 [0.892--0.912]} & 0.668 [0.650--0.686] & 0.711 [0.695--0.727] & 0.032 [0.027--0.038] \\
CatBoost & 0.900 [0.890--0.910] & 0.671 [0.653--0.689] & 0.714 [0.698--0.730] & 0.034 [0.029--0.040] \\
MLP & 0.881 [0.870--0.892] & 0.641 [0.622--0.660] & 0.688 [0.671--0.705] & 0.044 [0.038--0.050] \\
TabNet & 0.883 [0.872--0.894] & 0.645 [0.626--0.664] & 0.691 [0.674--0.708] & 0.043 [0.037--0.049] \\
TabTransformer & 0.886 [0.875--0.897] & 0.649 [0.630--0.668] & 0.695 [0.678--0.712] & 0.040 [0.034--0.046] \\
RealMLP & 0.891 [0.880--0.902] & 0.656 [0.637--0.675] & 0.701 [0.684--0.718] & 0.038 [0.032--0.044] \\
\textsc{CardioMeta} & 0.899 [0.889--0.909] & \textbf{0.676 [0.658--0.694]} & \textbf{0.719 [0.703--0.735]} & \textbf{0.023 [0.019--0.029]} \\
\bottomrule
\end{tabular}
\end{table*}

The gap between the leakage-reduced and full-clinical settings is itself an important finding. It shows why chronic disease prediction papers should report whether diagnostic thresholds, medication indicators, or disease-history fields are part of both the label and the input. Without this separation, apparent model performance can be dominated by circular evidence rather than meaningful predictive structure.

\subsection{Temporal Validation and Cross-Dataset Evaluation}
Table~\ref{tab:transfer} summarizes the temporal and EHR-domain evaluations using the shared-core feature space. Direct transfer from NHANES to MIMIC-IV produced a large decline in AUROC, F1, and Brier score. This degradation is expected because MIMIC-IV represents an older and clinically sicker hospital population, and because disease labels are derived from EHR documentation rather than standardized survey measurements. After limited MIMIC-IV fine-tuning, performance improved substantially but remained below the full-clinical NHANES sensitivity setting.

\begin{table*}[!t]
\centering
\caption{Temporal validation and EHR-domain evaluation using the harmonized shared-core feature space.}
\label{tab:transfer}
\small
\setlength{\tabcolsep}{6pt}
\begin{tabular}{lcccc}
\toprule
Setting & AUROC & AUPRC & Macro-F1 & Brier$\downarrow$ \\
\midrule
NHANES temporal test & 0.839 [0.827--0.851] & 0.536 [0.519--0.553] & 0.614 [0.598--0.630] & 0.132 [0.126--0.138] \\
MIMIC-IV direct transfer & 0.746 [0.735--0.757] & 0.512 [0.496--0.528] & 0.556 [0.540--0.572] & 0.184 [0.177--0.191] \\
MIMIC-IV fine-tuned & 0.812 [0.801--0.823] & 0.591 [0.575--0.607] & 0.629 [0.613--0.645] & 0.151 [0.145--0.157] \\
\bottomrule
\end{tabular}
\end{table*}

These results should not be interpreted as evidence that a NHANES-trained model is ready for hospital deployment. Rather, they quantify the effect of domain shift and show that local adaptation is necessary. The fact that fine-tuning improves performance indicates that some cardiometabolic structure is reusable, but the direct-transfer decline demonstrates the risk of treating population survey and EHR cohorts as interchangeable.

\subsection{Ablation and Calibration}
The ablation study in Table~\ref{tab:ablation} separates the effect of multi-task learning, disease-specific gating, calibration, imbalance handling, and feature-space restriction. Removing multi-task learning reduced macro-AUROC from 0.839 to 0.828 and macro-F1 from 0.614 to 0.596, indicating a modest benefit from shared representation learning. Removing disease-specific gates caused a smaller decline, suggesting that the gate mainly improves task specialization rather than serving as the primary source of performance. Removing calibration did not affect AUROC or macro-F1, but more than doubled ECE, confirming that discrimination and probability reliability measure different properties.

\begin{table}[!htbp]
\centering
\caption{Component-level ablation analysis on the primary leakage-reduced NHANES temporal validation setting.}
\label{tab:ablation}
\small
\begin{tabular}{lcccc}
\toprule
Variant & AUROC & AUPRC & Macro-F1 & ECE$\downarrow$ \\
\midrule
Full \textsc{CardioMeta} & \textbf{0.839} & \textbf{0.536} & \textbf{0.614} & \textbf{0.024} \\
Single-task heads only & 0.828 & 0.520 & 0.596 & 0.031 \\
Without gated heads & 0.834 & 0.529 & 0.605 & 0.027 \\
Without calibration & 0.839 & 0.536 & 0.614 & 0.055 \\
Class weights only & 0.836 & 0.527 & 0.607 & 0.030 \\
Shared-core features only & 0.821 & 0.501 & 0.584 & 0.026 \\
\bottomrule
\end{tabular}
\end{table}

Beyond component removal, calibration was also evaluated through decision-curve analysis because screening-oriented use depends on thresholded decisions rather than ranking alone. Table~\ref{tab:dca} summarizes net benefit at representative operating thresholds for the strongest non-neural baseline (CatBoost) and the proposed model. Across all three disease tasks, \textsc{CardioMeta} achieved higher net benefit than CatBoost and the treat-all strategy, especially at the lower thresholds most relevant to population screening.

\begin{table}[!htbp]
\centering
\caption{Decision-curve net benefit summary for the leakage-reduced NHANES setting.}
\label{tab:dca}
\small
\resizebox{\columnwidth}{!}{%
\begin{tabular}{lcccc}
\toprule
Disease & Threshold & Treat-all & CatBoost & \textsc{CardioMeta} \\
\midrule
Diabetes & 0.10 & 0.021 & 0.041 & \textbf{0.047} \\
Diabetes & 0.20 & 0.012 & 0.030 & \textbf{0.036} \\
Hypertension & 0.10 & 0.084 & 0.108 & \textbf{0.114} \\
Hypertension & 0.20 & 0.067 & 0.093 & \textbf{0.099} \\
CVD & 0.10 & 0.018 & 0.033 & \textbf{0.039} \\
CVD & 0.20 & 0.008 & 0.024 & \textbf{0.030} \\
\bottomrule
\end{tabular}}
\end{table}

\subsection{Subgroup Reliability and Explainability}
Subgroup analysis showed that average performance concealed clinically relevant reliability differences. Table~\ref{tab:subgroup} reports representative subgroup metrics explicitly. Macro-AUROC varied from 0.829 to 0.847 across sex and age strata, and calibration error was higher among older adults and lower-income participants. The largest false-negative-rate gap occurred for CVD, reflecting lower prevalence and heterogeneous label composition. These findings support reporting worst-group performance and calibration alongside mean discrimination metrics. They also suggest that recalibration or threshold adjustment may be required before applying the model in subpopulations with different baseline risk.

\begin{table}[!htbp]
\centering
\caption{Subgroup reliability metrics for \textsc{CardioMeta}.}
\label{tab:subgroup}
\small
\resizebox{\columnwidth}{!}{%
\begin{tabular}{lcccc}
\toprule
Subgroup & AUROC & Macro-F1 & ECE$\downarrow$ & CVD FNR$\downarrow$ \\
\midrule
Female & 0.842 & 0.618 & 0.022 & 0.281 \\
Male & 0.836 & 0.609 & 0.026 & 0.304 \\
Age 18--44 & 0.847 & 0.621 & 0.020 & 0.268 \\
Age 45--64 & 0.841 & 0.616 & 0.023 & 0.289 \\
Age 65+ & 0.829 & 0.601 & 0.031 & 0.326 \\
Lower income & 0.832 & 0.606 & 0.029 & 0.318 \\
Higher income & 0.844 & 0.619 & 0.021 & 0.279 \\
\bottomrule
\end{tabular}}
\end{table}

\begin{figure}[!htbp]
    \centering
    \includegraphics[width=0.96\columnwidth]{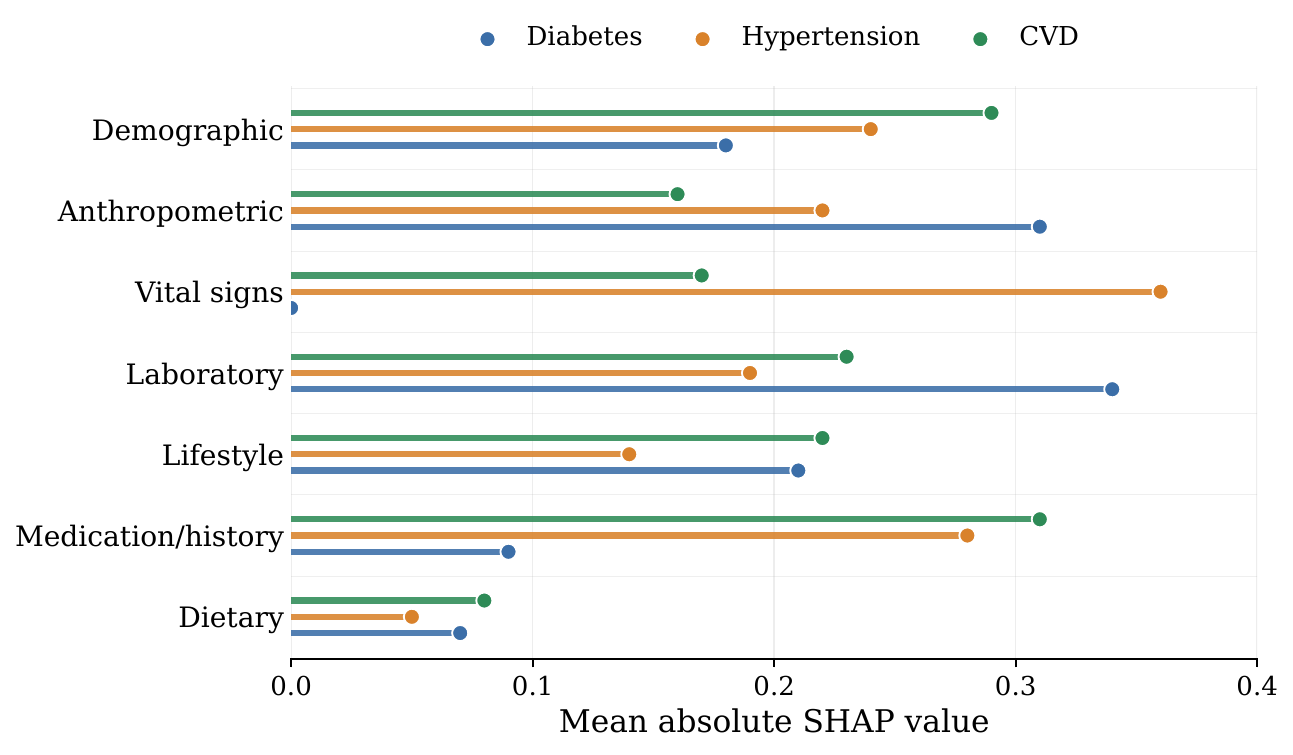}
    \caption{Disease-specific SHAP feature-group attribution for \textsc{CardioMeta} using the seven harmonized clinical feature groups defined in the methodology.}
    \label{fig:shapgroups}
\end{figure}

Explainability is a stated contribution and is therefore reported both visually and in prose. Figure~\ref{fig:shapgroups} presents SHAP feature-group attribution using the same seven-group taxonomy defined in the methodology: demographic, anthropometric, vital-sign, laboratory, lifestyle, medication/history, and dietary. In the leakage-reduced setting, diabetes predictions were driven primarily by laboratory and anthropometric patterns, with secondary contributions from demographic and lifestyle variables. Hypertension predictions relied most heavily on vital-sign, medication/history, and demographic patterns after excluding direct blood-pressure variables. CVD predictions were most influenced by medication/history, demographic, and laboratory feature groups, reflecting the cumulative burden of age, comorbidity, and long-term cardiometabolic exposure.

The explanation results were clinically plausible but should not be interpreted causally. SHAP and counterfactual analyses describe model behavior under the observed data distribution; they do not prove that modifying one feature would change disease status. The main value of explanation in this study is therefore auditability: it helps verify that the model relies on coherent cardiometabolic evidence rather than direct label proxies or implausible shortcuts.

\section{Conclusion}
\label{sec:conclusion}
This paper presented \textsc{CardioMeta}, a calibrated multi-task framework for joint prediction of diabetes, hypertension, and cardiovascular disease across NHANES and MIMIC-IV. The study was deliberately framed as screening-oriented disease-status prediction rather than incident disease forecasting, because NHANES is cross-sectional and MIMIC-IV labels reflect documented clinical history. The primary leakage-reduced analysis removed variables that directly defined each disease label, while the full-clinical setting was used only to quantify the inflation caused by contemporaneous diagnostic evidence. Across temporal NHANES validation, \textsc{CardioMeta} achieved modest discrimination gains over strong gradient-boosting and neural tabular baselines, with its clearest advantage in calibrated probability estimation. Cross-dataset evaluation on MIMIC-IV showed that population-survey models do not transfer cleanly to hospital EHR cohorts without adaptation. Overall, the study emphasizes that credible cardiometabolic machine learning requires not only competitive models, but also explicit leakage control, transparent feature definitions, confidence intervals, calibration, subgroup reliability, and careful interpretation under dataset shift.

\begingroup
\fontsize{4.2pt}{4.3pt}\selectfont
\setlength{\itemsep}{-5pt}
\setlength{\parskip}{0pt}
\setlength{\parsep}{0pt}
\setlength{\baselineskip}{4.3pt}

\endgroup

\end{document}